\documentclass[10pt,twocolumn,letterpaper]{article}

\usepackage{cvpr}
\usepackage{times}
\usepackage{epsfig}
\usepackage{graphicx}
\usepackage{amsmath}
\usepackage{amssymb}
\usepackage{epigraph}
\usepackage{pifont}
\usepackage{color}
\usepackage{xcolor}
\usepackage{amsthm}
\newtheorem{theorem}{Theorem}
\usepackage{amsfonts}
\usepackage{sidecap}
\usepackage{booktabs}
\usepackage{caption}
\usepackage{arydshln}
\usepackage{multirow}
\usepackage[export]{adjustbox}
\usepackage{diagbox}
\usepackage{colortbl}

\newcommand{\cmark}{\ding{51}}%
\newcommand{\xmark}{\ding{55}}%

\newcommand\blfootnote[1]{%
  \begingroup
  \renewcommand\thefootnote{}\footnote{#1}%
  \addtocounter{footnote}{-1}%
  \endgroup
}

\definecolor{sh_gray}{rgb}{0.84,0.84,0.84}
\definecolor{sh_gray2}{rgb}{1,0.89,0.75}
\definecolor{color3}{rgb}{0.95,0.95,0.95}
\definecolor{color4}{rgb}{0.96,0.96,0.86}

\newcommand{\best}[1]{\colorbox{sh_gray2}{\textbf{#1}}}%
\newcommand{\sbest}[1]{\colorbox{sh_gray}{\textbf{#1}}}%

\newcommand{\bx}{\mathbf{x}}
\newcommand{\bw}{\mathbf{w}}

\newcommand{\bff}{\mathbf{f}}
\newcommand{\parl}{{{\parallel}}}
\newcommand{\bu}{\mathbf{u}}

\usepackage[breaklinks=true,bookmarks=false]{hyperref}

\cvprfinalcopy

\def\cvprPaperID{2230}

\ifcvprfinal\fi

\title{\vspace{-0.7em}Striking the Right Balance with Uncertainty}

\begin{document}
\author{Salman Khan$^{*}$ \quad Munawar Hayat$^{*}$ \quad Syed Waqas Zamir \quad
Jianbing Shen$^{\dagger}$ \quad Ling Shao \\
Inception Institute of Artificial Intelligence,  UAE\\
{\tt\small {firstname.lastname}@inceptioniai.org}
}

\twocolumn[{
\renewcommand\twocolumn[1][]{#1}%
\maketitle
\begin{center}
	\centering
    \begin{minipage}[c]{0.63\textwidth}\vspace{-1em}
    \includegraphics[width=1\textwidth]{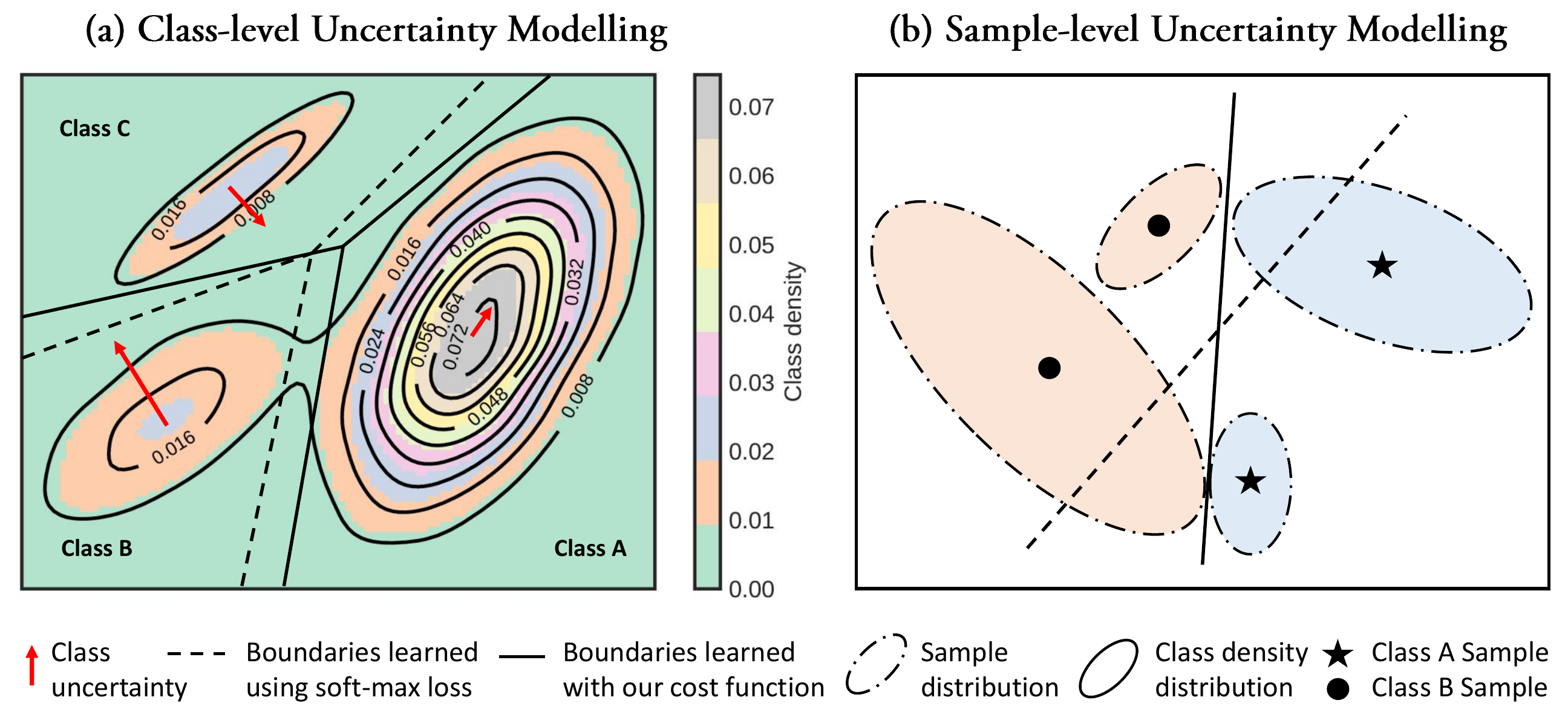}
    \end{minipage}\hfill
    \begin{minipage}[c]{0.36\textwidth}
        \captionof{figure}{Imbalance learning with Bayesian uncertainty estimates. \emph{(a)} We enforce class-level margin penalty based on class uncertainty. This pushes boundaries further away from rare classes B and C.  \emph{(b)} We also consider sample-level uncertainty that is modeled as a Gaussian distribution. The learned margins consider the confidence level of classifier to re-adjust boundaries (i.e., provide more room to uncertain samples). This improves the generalization ability of the proposed model for imbalanced classes. }
    \label{fig:intro}
    \end{minipage}
\end{center}%
}]


\begin{abstract}\vspace{-1.0em}
\blfootnote{$^*$Equal contribution, $^{\dagger}$Corresponding author}
   Learning unbiased models on imbalanced datasets is a significant challenge. Rare classes tend to get a concentrated representation in the classification space which hampers the generalization of learned boundaries to new test examples. In this paper, we demonstrate that the Bayesian uncertainty estimates directly correlate with the rarity of classes and the difficulty level of individual samples. Subsequently, we present a novel framework for uncertainty based class imbalance learning that follows two key insights: \textit{First}, classification boundaries should be extended further away from a more uncertain (rare) class to avoid over-fitting and enhance its generalization. \textit{Second}, each sample should be modeled as a multi-variate Gaussian distribution with a mean vector and a covariance matrix defined by the sample's uncertainty. The learned boundaries should respect not only the individual samples but also their distribution in the feature space. Our proposed approach efficiently utilizes sample and class uncertainty information to learn robust features and more generalizable classifiers. We systematically study the class imbalance problem and derive a novel loss formulation for max-margin learning based on Bayesian uncertainty measure. The proposed method shows significant performance improvements on six benchmark datasets for face verification, attribute prediction, digit/object classification and skin lesion detection. 
 \end{abstract}

\section{Introduction}
Objects, events, actions and visual concepts appear with varying frequencies in real world imagery \cite{rahman2018zero}. This often leads to highly skewed datasets where a few abundant classes outnumber several rare classes in a typical long-tail data distribution. The low amount of training data for infrequent classes makes it challenging to learn optimal classification boundaries in the feature space. Existing approaches to tackle class imbalance either modify data distribution \cite{saez2015smote,chawla2002smote,jeatrakul2010classification} or introduce appropriate costs to re-weight class errors \cite{khan2017cost, akbani2004applying, ren2018learning}. The popular data-level approaches are prone to over-fitting while the cost-sensitive learning requires careful choice of weights for successful training. Despite an overwhelming success of deep neural networks on computer vision problems, learning from highly imbalanced sets is still an open problem for deep learning \cite{khan2018guide}. 

This paper proposes a new direction towards learning balanced representations using deep neural networks (Fig.~\ref{fig:intro}). We use a principled approach to integrate Bayesian uncertainty estimates for class imbalance learning at two distinct levels, i.e., category-level and individual sample-level. Our approach is based on the observation that rare classes have higher uncertainty in the prediction space and the associated classifier confidence levels are low. Therefore, the uncertainty estimates can be used to expand decision regions for less frequent classes so that classifier's generalization to new test examples is improved. This concept is illustrated in Fig.~\ref{fig:intro}. Since all samples within a class do not have a uniform difficulty level, our approach also optimizes margins with respect to the uncertainty associated with individual samples.  The basic intuition for both cases is the same: a classifier should assign larger regions to more uncertain (rare) samples/classes.

\noindent \textbf{Related work:} State-of-the-art deep imbalance learning methods mainly propose novel objective functions \cite{hayat2019max}. Khan \etal \cite{khan2017cost} presented a cost-sensitive loss for CNNs where class-specific weights were automatically learned. Huang \etal \cite{huang2016learning} suggested a combination of triplet and quintuplet losses to preserve local class structures. A modified softmax was proposed in \cite{liu2017sphereface} to maximize the angular margin, thus avoiding  class imbalance. Quite recently, a meta-learning approach in \cite{ren2018learning} used selective instances for training on imbalanced sets. These methods have their respective limitations, e.g., \cite{khan2017cost} only considers class-level costs, \cite{huang2016learning} is not differentiable and requires heavy pre-processing for  quintuplet creation, \cite{liu2017sphereface} can only maximize margin on the hypersphere surface and \cite{ren2018learning} used an additional validation set to assign sample weights. Concurrent to this work, \cite{cui2019classbalancedloss} re-weights the loss by the inverse effective number of samples to learn balanced representations.

\noindent \textbf{Contributions:} Our approach is distinct in two ways: (a) this is the first work to link class imbalance with Bayesian uncertainty estimates \cite{gal2016dropout}, that have shown great promise on other tasks \cite{Kendall_2018_CVPR, feinman2017detecting, tzelepis2017linear},  and (b) we incorporate both class and sample-level confidence estimates to appropriately reshape learned boundaries. The paper therefore introduces the following major novelties. \textbf{(1)} A principled margin-enforcing formulation for softmax loss, underpinned by the Bayesian uncertainty estimates. \textbf{(2)} Sample modeling using multi-variate Gaussian distributions. The class boundaries are optimized to respect second order moments which improves generalization. \textbf{(3)} A fully differentiable loss formulation that can be easily plugged into existing architectures and used alongside other regularization techniques. 



\section{The Imbalance Problem}
\begin{figure}
\centering
\includegraphics[width=\columnwidth]{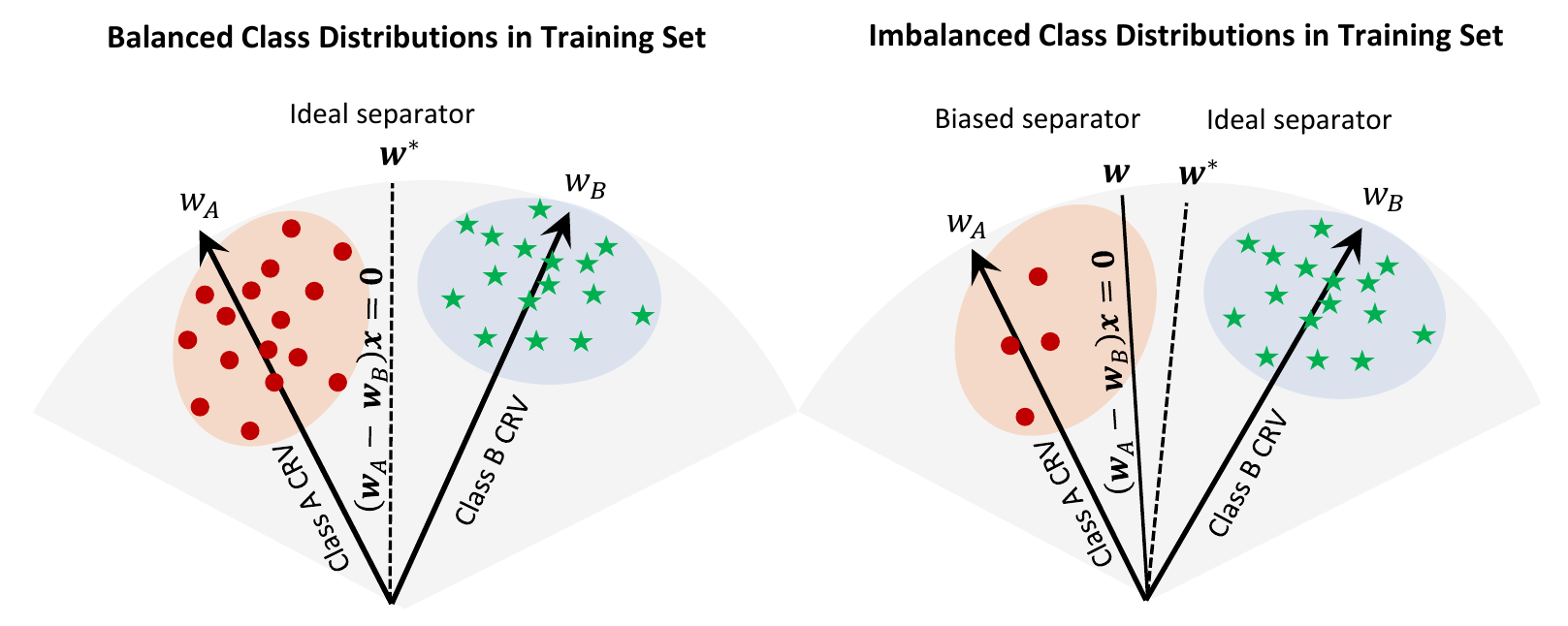}
\vspace{-2em}
\caption{Illustration for the class imbalance problem. True class distributions are shown in green and red. Unbalanced distributions lead to a skewed classification boundary that is biased towards the minority class. }
\label{fig:illus}
\end{figure}

We begin with an in depth analysis of the imbalance problem and draw several insights which lead to our proposed framework. We base our analysis around softmax loss, which is the most popular objective function for classification. For brevity, we consider a simplistic case of binary classification with two classes $A$ and $B$ in the training set denoted by image-label pairs: $\mathcal{D}= \{\bx^k, y^k\}_{k=1}^{K}$. The goal is to learn an optimal set of `representative vectors' ($\mathbf{w}_A, \mathbf{w}_B$ for classes $A$ and $B$, respectively) that lead to minimal empirical loss on set $\mathcal{D}$. The Class-Representative Vectors (CRV) define a loss-minimizing hyper-plane `$\mathbf{w}$', which is the boundary between two classes, i.e., given a feature projection `$\mathbf{f}$' corresponding to an input image $\mathbf{x}$, $\mathbf{f} \in \mathbf{w} \text{ iff } (\mathbf{w}_A - \mathbf{w}_B) \mathbf{f} = 0$ (ignoring unit biases). The class imbalance problem exists when the class frequencies $\tau_A, \tau_B$ are greatly mismatched in the set $\mathcal{D}$. As illustrated in Fig.~\ref{fig:illus}, in such cases, the hypothesis ($\mathbf{w}$) learned on $\mathcal{D}$ using a softmax loss can be biased towards the minority class and significantly different from the ideal separator ($\mathbf{w}^*$). Next, we breakdown the imbalance problem and explain underlying reasons. 

\subsection{Bias due to Empirical Loss Minimization}
We consider $\mathbf{w}^*$ to be an optimal boundary obtained by loss minimization with respect to the actual hidden distributions $P_A$ and $P_B$ of classes, i.e.:
\begin{align}\small
\mathbf{w}^* &= \arg\min_{\mathbf{w}} \mathcal{L}_P(\mathbf{w}), \text{ where, }\notag\\
 \mathcal{L}_P(\mathbf{w})&= \int_{\mathcal{R}_B^{\mathbf{w}}} P_A(\mathbf{f})d\mathbf{f} + \int_{\mathcal{R}_A^{\mathbf{w}}} P_B(\mathbf{f})d\mathbf{f},
\end{align}
and $\mathcal{R}^{\mathbf{w}}_A, \mathcal{R}^{\mathbf{w}}_B$ denote the classification regions for classes $A$ and $B$, respectively. Given $\mathcal{D}$, the empirical loss calculated on the training set is:
\begin{align}\small
\mathcal{L}_{\mathcal{D}}(\mathbf{w}) = &\#\{\mathbf{x}^k|\mathbf{f}^k \in \mathcal{D}^{A} \wedge \mathbf{f}^k  \in \mathcal{R}_B^{\mathbf{w}} \} +  \notag\\ 
& \#\{\mathbf{x}^k|\mathbf{f}^k \in \mathcal{D}^{B} \wedge \mathbf{f}^k  \in \mathcal{R}_A^{\mathbf{w}} \}
\end{align}
Further, assume that the normalized class frequencies $\tau_A$ and $\tau_B$ are related as $\tau_A + \tau_B = 1$. Then, the expected empirical loss for any hypothesis $\mathbf{w}$ is: 
\begin{align}\label{eq:empirical}\small
\mathbb{E}[\mathcal{L_{\mathcal{D}}(\mathbf{w})}] = \tau_A \int_{\mathcal{R}_B^{\mathbf{w}}} P_A(\mathbf{f})d\mathbf{f} + \tau_B \int_{\mathcal{R}_A^{\mathbf{w}}} P_B(\mathbf{f})d\mathbf{f}.
\end{align}
Note that $\tau_A \neq \tau_B \neq 0.5$ due to class imbalance and typically $|\tau_A - \tau_B | > 0.5$ in practical cases where a significant imbalance ratio exists. Next, we show that when large imbalance exists, the learned classification boundaries are biased towards minority classes. 

\begin{theorem}
For high imbalanced ratios, minimization of empirical loss results in a hypothesis $\hat{\mathbf{w}}$ that is highly likely to be biased towards the minority class `$z$' such that $\mathcal{R}_{z}^{\mathbf{w}^*} > \mathcal{R}_{z}^{\hat{\mathbf{w}}}$. In other words, the classification region induced by the optimal separator is larger than the one induced by empirically learned boundary.
\end{theorem}
\begin{proof}
According to Eq.~\ref{eq:empirical}, due to the imbalanced proportion among classes, $\mathbf{w}^*$ is 
more likely to incur higher empirical error than an alternate hypothesis based on an empirical loss, i.e., for any $\hat{\mathbf{w}}: \mathcal{R}^{z}_{\mathbf{w}^*} {>} \mathcal{R}^{z}_{\hat{\mathbf{w}}}$, it is more likely that: 
\begin{align}\label{eq:loss_relation}\small
&\mathcal{L}_{\mathcal{D}}(\mathbf{w}^*) > \mathcal{L}_{\mathcal{D}}(\hat{\mathbf{w}}) \text{ because } 
 \tau_A \int_{\mathcal{R}_B^{\mathbf{w}^*}} P_A(\mathbf{f})d\mathbf{f} + \notag\\
&  \tau_B \int_{\mathcal{R}_A^{\mathbf{w}^*}} P_B(\mathbf{f})d\mathbf{f} >  \tau_A \int_{\mathcal{R}_B^{\hat{\mathbf{w}}}} P_A(\mathbf{f})d\mathbf{f} 
+ \tau_B \int_{\mathcal{R}_A^{\hat{\mathbf{w}}}} P_B(\mathbf{f})d\mathbf{f}.
\end{align}
Then, for a significant imbalance ratio such that $\tau_z << \tau_{\text{rest}}$, it directly follows that $\mathcal{R}_{z}^{\mathbf{w}^*} > \mathcal{R}_{z}^{\hat{\mathbf{w}}}$. Intuitively, this is a natural implication of imbalanced class  distribution which forces the classifier to shift $\hat{\mathbf{w}}$ closer to minority classes because it reduces empirical error. The likelihood of classifier bias is  directly proportional to the imbalance rate. 
\end{proof}
A common strategy to tackle data imbalance is through the introduction of cost-sensitive loss functions \cite{khan2017cost}. We briefly elaborate on the effect of these losses next and explain why this solution is sub-optimal.

\subsection{Cost-sensitive Loss}
From Eq.~\ref{eq:loss_relation}, one simple solution seems to be the introduction of costs to re-weight the minority class errors. Existing cost-sensitive losses (particularly those based on deep-networks \cite{khan2017cost}) adopt this idea and assign score-level penalty to the minority class predictions. This means that the classifier is forced to correctly classify training samples belonging to minority classes. This approach has certain limitations. \textbf{(1)} Appropriately tuning class specific costs is a challenging task as it requires domain-knowledge with costs usually fixed at the beginning and not dynamically changed during the course of training. \textbf{(2)} A more stringent caveat is that such costs do not affect the learned boundaries $\hat{\mathbf{w}}$ if the training samples are separable \cite{wallace2011class,khan2017cost}. Further, when the classes are non-separable, minority class representation in the dataset is directly proportional to its mis-classification probability, i.e.,: $\tau_z \propto \int_{\mathcal{R}_{rest}^{\hat{\mathbf{w}}}} P_z(\mathbf{f})d\mathbf{f}$. 
\textbf{(3)} Generally, these costs are not applied at test time and therefore class-boundaries are effectively unchanged. In summary, while this practice enforces the classifier to more accurately classify training samples belonging to minority classes,  it does not enhance generalization capability of the learned model. This can be understood from the relation for generalization error and  Fig.~\ref{fig:illus}. For an empirical distribution $Q$ for classes $A$ and $B$, the generalization error (expected loss on the test set $\mathcal{T}$) is:
\begin{align}\label{eq:gen_error}\small
\mathbb{E}[\mathcal{L_{\mathcal{T}}(\mathbf{w})}] = \tau'_A \int_{\mathcal{R}_B^{\mathbf{w}}} Q_A(\mathbf{f})d\mathbf{f} + \tau'_B \int_{\mathcal{R}_A^{\mathbf{w}}} Q_B(\mathbf{f})d\mathbf{f},
\end{align}
where $\tau'_A, \tau'_B$ are the normalized frequencies on the test set.
The paper aims to overcome these existing limitations by proposing a new loss formulation that seeks to simultaneously extend minority class boundaries and enforce margin constraints on less represented classes to achieve better generalization performance. We provide details of our technique in the next section. 

\begin{figure}[t]
\includegraphics[width=\columnwidth]{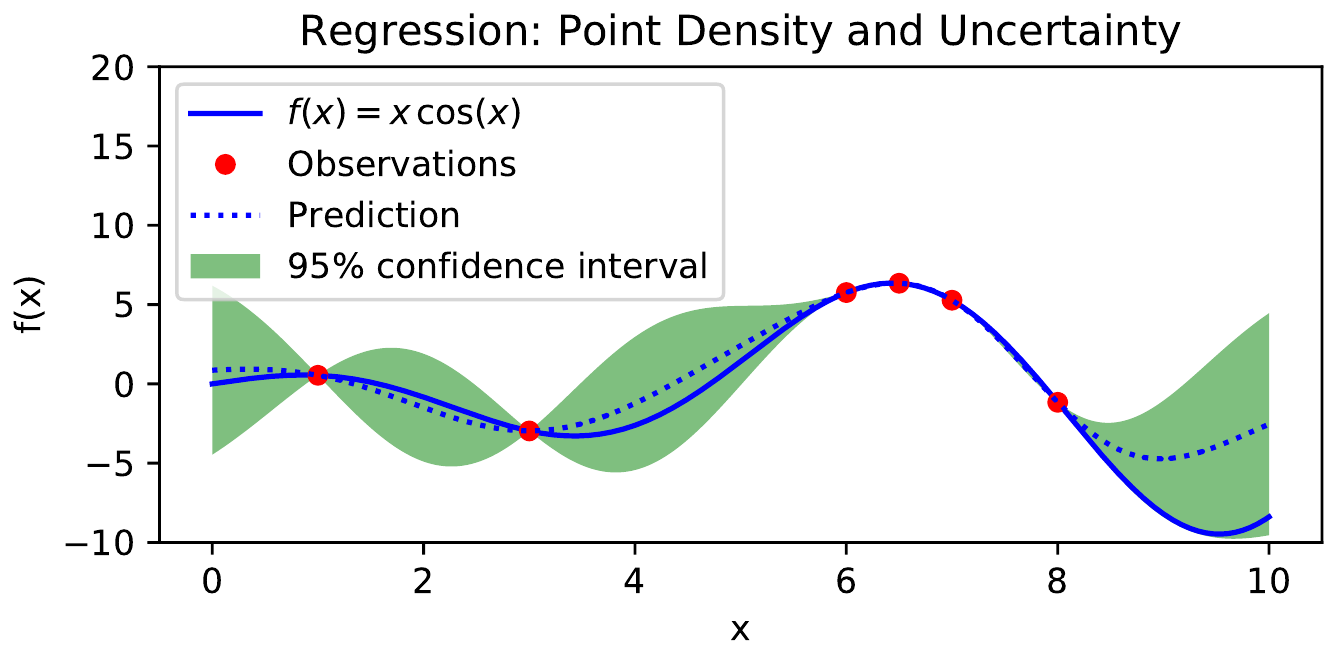}
\includegraphics[width=0.95\columnwidth,right]{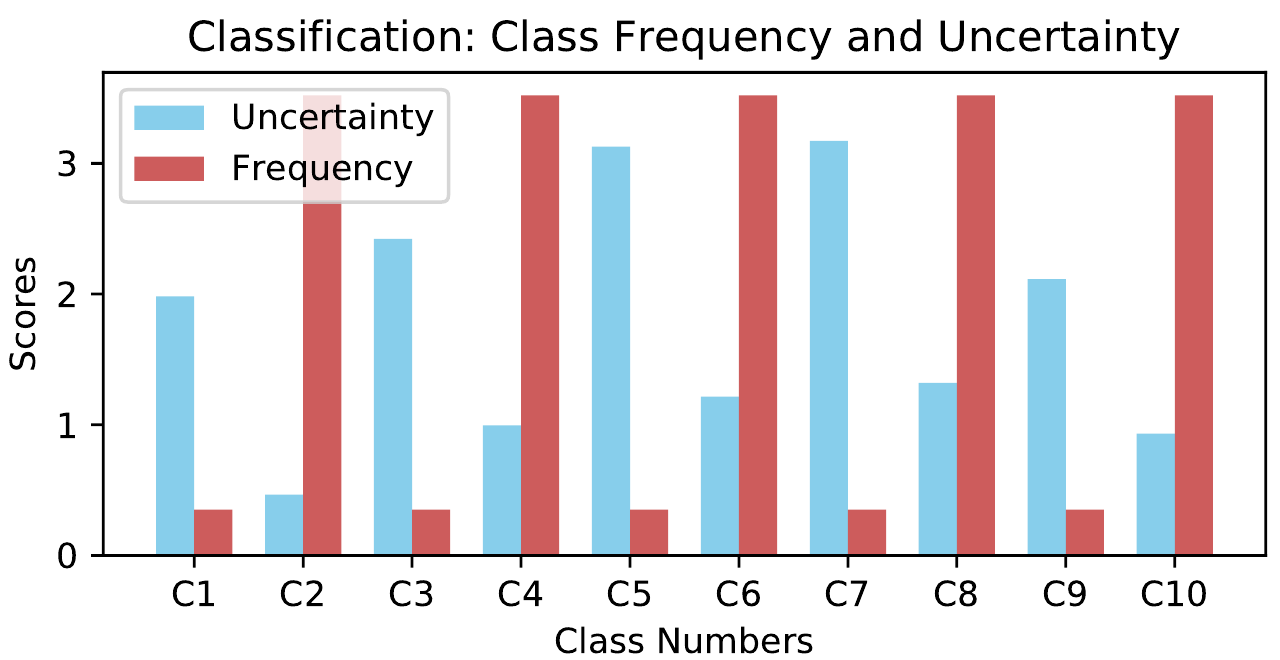}
\caption{\emph{Top:} One-dimensional Gaussian process regression using maximul-likelihood estimation. The lack of observations results in higher confidence levels. \emph{Bottom:} The uncertainty estimates for imbalanced CIFAR-10 dataset. The uncertainty is higher for classes with less representation. }
\label{fig:reg_cls}
\end{figure}

\section{Bayesian Uncertainty Estimates}
Bayesian models can provide uncertainty estimates alongside output predictions. Given an input, the uncertainty estimates correspond to the confidence level for each outcome predicted by the model. We hypothesize that the confidence-level of predictions is directly related to the class representation in the training set. As illustrated in Fig.~\ref{fig:reg_cls}, under-represented classes in the training set lead to higher uncertainty and bigger confidence intervals. In contrast, well-represented classes are associated with less uncertainty and  compact confidence intervals.

We use deep CNNs with dropout to obtain Bayesian uncertainty estimates. It has been proved that dropout-based deep networks provide an approximation to Gaussian process \cite{gal2016dropout}. A Gaussian process is a Bayesian technique because it constructs a prior distribution over a family of functions $\mathcal{F}$ \cite{rasmussen2004gaussian}. This distribution is updated conditioned on observations, i.e., all the functions that are consistent with the labels are retained. At inference time, an output is obtained from each of the functions and expectation is computed to generate the final prediction. The variance of these outputs gives an uncertainty estimate. In the following, we first provide an overview of dropout and then describe uncertainty computation using dropout.

\textbf{Dropout:} Dropout was originally proposed as a regularization measure for deep neural networks \cite{srivastava2014dropout}.  During training, a sub-network is sampled from the full network by randomly dropping a set of neurons. In this manner, each neuron is activated with a fixed probability `$p$'. At test time, full model is used for prediction and the output activations are multiplied with the probability $p$ to obtain expectation. Suppose the network parameters are denoted by $\Theta = \{\theta_1, \ldots, \theta_L\}$ for a total of $L$ network layers. Then, by applying masks $\mathbf{m}$ generated using i.i.d binary distributions, we can  obtain $N$ samples all corresponding to different network configurations $\hat{\Theta}$ that form an ensemble $\mathcal{M}$: 
\begin{align}
\mathcal{M} =& \{\hat{\Theta}^{i}: i \in [1,N]\}, \;\text{where, } \hat{\Theta}^i = \Theta \circ \mathbf{m}^i, \notag\\
\mathbf{m}^i =&  \{m_{l} : l \in [1,L]\}, \; s.t., m_{l} \sim \text{Bernoulli}(p)
\end{align} 

\textbf{Uncertainty:}
For each input $\mathbf{x}_k$, $N$ model configurations are applied to obtain a set of outputs $\{\hat{\mathbf{y}}\}$. The expected output is calculated using Monte Carlo estimate for first moment ($\mathbb{E}_{q(\mathbf{y}|\mathbf{x})}[\mathbf{y}]$): $\small
\mathbf{y}\approx \frac{1}{N} \sum_{i=1}^{N} \hat{\mathbf{y}}(\mathbf{x}; \hat{\Theta}^i),$
where $q$ denotes an output distribution that approximates the intractable posterior distribution of deep Gaussian process. The uncertainty is calculated using the second moment ($\mathbb{V}_{q(\mathbf{y}|\mathbf{x})}[\mathbf{y}]$) through Monte Carlo estimation:
\begin{align}\small
\bu  \approx \tau^{-1}\mathbf{I}_D + \frac{1}{N}\sum_{i=1}^{N} \hat{\mathbf{y}}^T \hat{\mathbf{y}} - 
\mathbb{E}_{q(\mathbf{y}|\mathbf{x})}[\mathbf{y}]^T \mathbb{E}_{q(\mathbf{y}|\mathbf{x})}[\mathbf{y}] , 
\end{align}
where $\tau$ is the  model precision (a function of weight decay) and $\mathbf{I}_C \in \mathbb{R}^{C\times C}$ is an identity matrix where $C$ denotes the number of classes.

\section{Uncertainty based Max-margin Learning}
The softmax loss can be computed for a given feature $\mathbf{f}$ and its true class label $y$ as follows:
\begin{align} \label{eq:softmax}
\mathcal{L}_{sm} = -\log \Big( \frac{\exp(\mathbf{w}^T_{y}\mathbf{f})}{\sum_j \exp( \mathbf{w}^T_{j}\mathbf{f})} \Big),
\end{align}
where $j \in [1,C]$ ($C$ is the number of classes). In the above loss formulation, we include the last fully connected layer within softmax loss which will be useful for our analysis later on. Further, for the sake of brevity, we do not mention bias in Eq.~\ref{eq:softmax}. Note that the dot-product $\mathbf{w}^T_{y}\mathbf{f}$ can also be expressed as $\bw^T_{y}\bff =  \parl \bw_{y} \parl\parl\bff\parl\cos(\alpha_y)$. Therefore, if a class $z$ is rare in the training set, for an input feature belonging to this class, the softmax loss enforces: 
\begin{align}\label{eq:sm_large}
\parl\bw_{z}\parl \parl\bff\parl \cos(\alpha_z)> \parl\bw_{rest}\parl \parl\bff \parallel\cos(\alpha_{rest}).
\end{align}
Intuitively, we would like to impose a large margin on more uncertain classes. Our experiments show that the class uncertainty is inversely proportional to its frequency in the training set, i.e., rare classes are more uncertain (Fig.~\ref{fig:reg_cls}). To improve generalization performance, we can impose a more strict constraint for uncertain classes:
$$\parl\bw_{z}\parl \parl\bff\parl \cos(m_z\alpha_z)> \parl\bw_{rest}\parl \parl\bff \parl\cos(m_{rest}\alpha_{rest}),$$
where, $m = \max(1,\lfloor 0.5\bu_{y}\rfloor$), $\bu_z > \bu_{rest}$, $0 \leq \alpha_z \leq \frac{\pi}{\bu_z}$ and $\bu_z \in \mathbb{R}^{+}$. This implies that the classifier will try to separate rare classes by a more rigorous margin. The margin maximizing softmax loss \cite{liu2016large} is defined as: 
\begin{align}\label{eq:mod_loss}
\mathcal{L}'_{sm} = -\log \Big( \frac{\exp(\parl\bw_y \parl \parl \bff \parl \psi(\alpha_y) )}{ \sum_{j} \exp(\parl\bw_j \parl \parl \bff \parl \psi(\alpha_j) ) }\Big),
\end{align}
where $\psi(\cdot)$ is a continuous and monotonically decreasing function in the range $[0,\pi]$:
\begin{align}
\psi(\alpha_j) = 
	\begin{cases} 
      (-1)^r \cos(m \alpha_j) {-} 2r & \alpha_{j = y} \in \Big[\frac{r\pi}{m}, \frac{(r+1)\pi}{m} \Big] \\
      \cos(\alpha_j) & j \neq y, 
   \end{cases}\notag
\end{align}
where $r \in [0,m{-}1]$ is an integer. The gradient back-propagation requires relations in terms of $\bw$ and $\bff$, therefore we substitute $\cos(m\alpha_j)$ with its expansion in terms of Chebyshev polynomials of the first kind ($T_m$), i.e.,
\begin{align}\small
\cos(m\alpha_j)  =  \sum_{t=0}^{\lfloor m/2 \rfloor} \Big(\begin{array}{c} m\\2t\end{array}\Big) (\cos^2(\alpha_j) - 1)^{m} \cos(\alpha_j)^{m-2t}. \notag
\end{align}
Here, $\cos(\alpha_j)$ is substituted with $\frac{\bw_j^T\bff}{\parl \bw_j\parl \parl \bff\parl}$. This gives us the max-margin formulation in terms of differentiable relations.


\section{Sample-level Uncertainty Modeling}
Although uncertainty driven class-level margin enforcement is important, not all samples in a class have equal difficulty level. The samples that can potentially be misclassified have larger uncertainties. We therefore propose a mechanism to incorporate sample-level uncertainty for imbalanced learning.
Existing classification networks only use the mean representation from a distribution of samples to represent each training example. Inspired by \cite{tzelepis2018maximum}, we propose to represent a single sample as a function of its first and second order moments. To this end, consider that the deep feature representations of input media is randomly sampled from a multi-variate Gaussian distribution: $\mathbf{f} \sim \mathcal{N}(\mathbf{\mu}_f, \Sigma_f)$, where $\mathbf{\mu}_f$ and $\Sigma_f$, respectively denote mean and covariance of the features. The softmax loss can be computed for a given feature $\mathbf{f}$ and its true class label $y$ using Eq.~\ref{eq:softmax}.

For an input feature to be correctly classified, its projection on the true class vector should give a maximum response (Eq.~\ref{eq:sm_large}). In contrast, an example is classified into a wrong category $k$ if the following condition holds true:
\begin{equation}
\exists j \in [1,C] \; s.t., \;\bw^T_{y}\bff < \bw^T_{j}\bff, \quad j \neq y.
\end{equation}
We are interested in quantifying the probability of misclassification taking into account the distribution of each sample. It can provide a measure of confidence for loss estimates computed on the training samples. Direct computation of the misclassification probability using softmax loss in Eq.~\ref{eq:softmax} is intractable and can only be approximated.  Therefore, we introduce a simpler error function that models the essential loss behavior.
\begin{align}
\mathcal{E}(\mathbf{f}) =  \bw^T_{j}\bff -  \bw^T_{y}\bff , \quad j \neq y.
\end{align}
 The formulation can be used to exactly compute the misclassification probability as we will show next. Since the error function is a linear transformation of input feature $\mathbf{f} \sim  \mathcal{N}(\mathbf{\mu}_f, \Sigma_f)$, the resulting error distribution is also a uni-variate Gaussian variable. The first and second order statistics of the error distribution can be given in terms of $\mathbf{\mu}_f, \Sigma_f$ as follows:
 \begin{align}
 \mu_{\mathcal{E}}  & = \mathbb{E}[\mathcal{E}(\mathbf{f})] =  (\mathbf{w}_{j} -  \mathbf{w}_{y})^T\mathbf{\mu}_f \notag\\
 \sigma_{\mathcal{E}}^2 &= \mathbb{E}[(\mathcal{E}(\mathbf{f})  -  \mu_{\mathcal{E}})^2] = (\mathbf{w}_{j} {-}  \mathbf{w}_{y})^T\Sigma_{f}(\mathbf{w}_{j} {-}  \mathbf{w}_{y})
 \end{align}
Now, the misclassification probability for a feature $\mathbf{f}$ can be linked with error distribution because $\mathcal{E}(\mathbf{f})> 0$ denotes a misclassification. The complementary cumulative probability distribution function (CCDF) $\hat{\mathbb{F}}_{\mathcal{E}}$ is given as follows:
\begin{align}
\hat{\mathbb{F}}_{\mathcal{E}}(0) = & \mathbb{P}(\mathcal{E}(\mathbf{f})> 0) = 1-   \mathbb{P}(\mathcal{E}(\mathbf{f})< 0) \notag\\
= & 1 - \mathbb{P}\Big( \frac{\mathcal{E}(\mathbf{f}) -  \mu_{\mathcal{E}} }{ \sigma_{\mathcal{E}} } < -\frac{\mu_{\mathcal{E}} }{\sigma_{\mathcal{E}}}\Big) \notag\\
\text{if } \mathbf{z} =  &  \frac{\mathcal{E}(\mathbf{f}) -  \mu_{\mathcal{E}} }{ \sigma_{\mathcal{E}} } \sim \mathcal{N}(0,1) \notag\\
\hat{\mathbb{F}}_{\mathcal{E}}(0) = & 1 - \mathbb{F}_{\mathbf{z}}(-\frac{\mu_{\mathcal{E}} }{\sigma_{\mathcal{E}}}) = \frac{1}{2} \Big( 1 + \text{erf}\Big[\frac{\mu_{\mathcal{E}}}{\sqrt{2\sigma^2_{\mathcal{E}}}}\Big]\Big),
\end{align}
where $\mathbb{F}_{\mathbf{z}}$ denotes the cumulative probability distribution function (CDF). The probability estimates are then used to re-weight the loss values such that uncertainty is incorporated. The function $\psi(\cdot)$ is modified as follows to obtain an improved loss function in Eq.~\ref{eq:mod_loss}:
\begin{align}\small
\psi(\alpha_j) = 
	\begin{cases} 
      \hat{\mathbb{F}}_{\mathcal{E}}(0)((-1)^r \cos(m \alpha_j) - 2r) & \alpha_{j=y} \in \Big[\frac{r\pi}{m}, \frac{(r+1)\pi}{m} \Big] \\
      \cos(\alpha_j) & j \neq y 
   \end{cases}
\end{align}
The loss defined above enforces a margin ($m$) between output predictions weighted by the cumulative probability ($\hat{\mathbb{F}}_{\mathcal{E}}(0)$). A higher uncertainty means a stricter margin based penalty for a class $j$.
The modified loss function becomes equal to the original softmax loss when the uncertainty is zero and $m=1$.


\section{Experiments}
\subsection{Datasets}

 \noindent \textbf{Face Recognition:} 
Facial recognition datasets commonly exhibit large-imbalance which poses a significant challenge for classifier learning. Following \cite{deng2018arcface}, we use VGG2 dataset \cite{Cao18} with 3,141,890 images of 8,631 subjects to train our deep network. We evaluate the trained model on four large-scale datasets namely Labeled Faces in the Wild (LFW) \cite{LFWTechUpdate} and YouTube Faces (YTF) \cite{wolf2011face}, AgeDB \cite{AgeDB} and Celebrities in Frontal Profile (CFP) \cite{cfp-paper}. 
LFW \cite{LFWTechUpdate} contains 13,233 web-collected images belonging to 5,749 different identities, with large variations in pose, expression and illumination. We follow the standard protocol of `unrestricted with labeled outside data'. 
 YTF \cite{wolf2011face} has 3,425 sequences of 195 subjects. We follow the standard evaluation protocol on 5,000 video pairs.  AgeDB \cite{AgeDB} dataset has 12,240 images of 440 subjects. The test set is divided into four groups with different year gaps (5, 10, 20 and 30 years). We only report the performance on the most challenging subset, AgeDB-30. CFP \cite{cfp-paper} has 500 subjects in total, each with 10 frontal and 4 profile images. In this paper, we only evaluate on the most challenging subset CFP-FP.

\noindent  \textbf{Skin Lesion Classification:} 
 Edinburgh Dermofit Image Library (DIL) consists of 1,300 high quality skin lesion images based on diagnosis from dermatologists and dermatopathologists. There are 10 types of lesions identified in this dataset including melanomas, seborrhoeic keratosis and basal cell carcinomas. The number of images in each category varies between 24 and 331 (mean 130, median 83). Similar to \cite{ballerini2013color}, we report results with 3-fold cross validation.

\noindent  \textbf{Digit/Object Classification:} We evaluate on imbalanced MNIST and CIFAR-10 datasets for generic digit and object classification. Standard MNIST consists of 70,000 images of handwritten digits (0-9). Out of these, 60,000 images are used for training (∼600/class) and the remaining 10,000 for testing (∼100/class). CIFAR-10 contains 60,000 images belonging to 10 classes (6,000 images/class). The standard train/test split for each class is $\sim$83.3\%/16.7\% images. We evaluate our approach on the standard split as well as on an artificially created imbalanced split. To imbalance the training distribution, we randomly drop $90\%$ of the samples for half of the classes.
 
\noindent  \textbf{Attribute Prediction:} We use the large-scale CelebA dataset \cite{liu2015deep} for (multi-label) facial attribute prediction task. This dataset consists of 202,599 images belonging to 10,177 human identities. Each image is annotated with a diverse set of 40 binary attributes. There exists a significant imbalance in the training set with ratios up to 1:43. Following the standard protocol \cite{liu2015deep}, we use 152,770 images for training, 10,000 for validation, and remaining 19,867 for testing. 
For evaluation, we report \textit{Balanced Classification Accuracy} (BCA) defined as:
$\textit{BCA}=0.5 {\times} \frac{t_p}{N_p} + 0.5 {\times} \frac{t_n}{N_n}$,
where $t_p$ and $t_n$, respectively denote true positives and true negatives, and $N_p$ and $N_n$ are total number of positive and negative samples. This evaluation metric is more suitable for multi-label imbalanced learning tasks since it gives equal weight to both majority and minority classes. Other evaluation metrics used in the literature \cite{liu2015deep} which define accuracy as $\frac{t_p+t_n}{N_p+N_n}$ can be biased towards majority classes.

\subsubsection{Implementation Details}
The uncertainty estimates are applied progressively during training. We start with standard softmax ($m=1$), followed by class-level uncertainty based max-margin learning and finally sample-level uncertainty modeling during the last 10 epochs. The compute intensive sample-level uncertainty estimates are therefore only done for few epochs. The proposed strategy can be related with curriculum learning, since it starts with a \textit{simple} task by considering a balanced class distribution, and gradually introduces \textit{harder} tasks by expanding or shrinking classification boundaries of different classes based upon their uncertainty estimates. For attribute prediction on CelebA dataset, the training times required for standard softmax and ours are $\sim$3.4 and 4.6 hours, respectively, on a Dell Precision 7920 machine with TitanXp GPU. In our experiments, we fixed $m=3$ since it gives relatively better results. Experiments on imbalanced CIFAR-10 for $m=2,3,4$ achieve an accuracy of $80.2\%,80.6\%,80.5\%$, respectively.  Values of $N \geq 5$ give stable uncertainty estimates. We fixed $N=10$ for the optimal trade-off between reliable uncertainty and compute efficiency. An ablation study on different values of $N=5,10,20,40$ on imbalanced CIFAR-10 results in respective accuracies of $80.4\%, 80.6\%, 80.6\%, 80.7\%$.

For Skin Lesion detection, we deploy ResNet-18 backbone with two fully connected layers (with intermediate rectified linear units non-linearities and dropout) inserted after the global pooling layer. For face verification tasks, we train  Squeeze and Excitation (SE) networks \cite{hu2017squeeze} with ResNet-50 backbone. The face images are pre-processed to $112\times112$ using multi-task cascaded CNN \cite{zhang2016joint}. After the network is trained on VGG-2 dataset, we use features extracted after global pooling layer for face verification evaluations. On imbalanced MNIST dataset, we use the same settings as in \cite{khan2017cost} to enable direct comparison with the recently proposed imbalanced learning technique \cite{khan2017cost}. For experiments on imbalanced CIFAR-10 dataset, we extract features from VGG16 \cite{Simonyan2015} pre-trained on ImageNet \cite{Deng2009}. A simple neural network with two hidden layers (512 neurons each) with dropouts is trained on the extracted features to get uncertainty estimates and perform classification. Training a network on VGG extracted features enables us to compare against traditional  imbalanced learning techniques. Specifically, data level under-sampling \& over-sampling methods in Table~\ref{res: cifar10} are used on VGG features, followed by the two layered NN for classification. For attribute prediction on CelebA dataset, we train a model with ResNet-50 backbone and two fully connected layers with dropout inserted after the global pooling and the first fully connected layers. The model is trained to minimize sum of binary cross entropy losses, using relatively smaller learning rates for the layers before the global pooling layer and larger rates for the layers inserted afterwards. 

\subsection{Results and Comparisons}

\begin{table}[]
\centering
 \resizebox{0.99\columnwidth}{!}{
\begin{tabular}{l l  c c}
\toprule[0.4mm]
\rowcolor{color3}
 & \textbf{Methods}    &  \textbf{LFW}             & \textbf{YTF} \\
  \midrule
 \multirow{8}{*}{\rotatebox[origin=c]{0}{\parbox[c]{2.5cm}{\centering Methods using non-public data }}}  & DeepFace \cite{taigman2014deepface}   & 97.35 & 91.4           \\
 & FaceNet \cite{schroff2015facenet}      & 99.63 & 95.4           \\
 & Web-scale \cite{taigman2015web}          & 98.37 & -              \\

 & DeepID2+ \cite{sun2015deeply}          & 99.47 & 93.2           \\
 & Baidu  \cite{liu2015targeting}         & 99.13 & -              \\
 & Center Face \cite{wen2016discriminative} & 99.28 & 94.9           \\
 & Marginal Loss \cite{deng2017marginal}  & 99.48 & 96.0          \\
 & Noisy Softmax \cite{chen2017noisy}    & 99.18  & 94.9          \\
\midrule
 \multirow{6}{*}{\rotatebox[origin=c]{0}{\parbox[c]{2.5cm}{\centering Novel Loss Functions}}}  & Softmax+Contrastive \cite{sun2014deep}    & 98.78  & 93.5           \\
 & Triplet Loss  \cite{schroff2015facenet}  & 98.70  & 93.4           \\
 & Large-Margin Softmax  \cite{liu2016large}  & 99.10 & 94.0 \\
 & Center Loss  \cite{wen2016discriminative}    & 99.05  & 94.4   \\
 & SphereFace  \cite{liu2017sphereface}        & 99.42  & 95.0  \\
 & CosFace  \cite{Wang_2018_CVPR}   & 99.33    & \sbest{96.1}           \\
 \midrule 
 \multirow{4}{*}{\rotatebox[origin=c]{0}{\parbox[c]{2.5cm}{\centering Imbalance Learning}}}   & Range Loss  \cite{zhang2017range}    & 99.52 & 93.7           \\
 & Augmentation \cite{masi2016we}       & 98.06  & -              \\
 & Center Inv. Loss   \cite{wu2017deep}   & 99.12 & 93.9   \\
 & Feature transfer  \cite{yin2018feature}     & 99.37 & -   \\
 & LMLE  \cite{huang2016learning}   & \sbest{99.51}    & {95.8}          \\
\midrule
 & This Paper  & \best{99.71}    &  \best{97.3}             \\
\bottomrule[0.4mm]
\end{tabular}}\vspace{-0.5em}
\caption{Face Verification Performance on LFW and YTF datasets. We trained our model on VGG2 dataset. Most
methods in the first cell use large-scale outside data that are not publicly available. The second cell includes
novel loss functions. The state-of-the-art imbalanced learning methods are in the last group.}
\label{tab:lfw_ytf}
\end{table}

\noindent \textbf{Face Verification:} We compare our approach with 20 recent and top-performing methods on LFW and YTF datasets (Table~\ref{tab:lfw_ytf}). We divide these methods into three categories: (a) methods that use large amounts of non-publicly available data sources to train their models, (b) methods that design novel loss functions for face verification and (c) methods that deal with data imbalance. We note that the performances on both LFW and YTF are currently saturated with many recent methods already surpassing human performance. Our method achieves competitive performance on both these datasets. Note that some of the compared methods used as much as 200M images \cite{schroff2015facenet} and an ensemble of 25 models \cite{sun2015deeply} for training. Further evaluations on additional datasets show that the proposed method achieves verification accuracies of $97.0\%$ and $94.4\%$ on CFP-FP and AgeDB-30 datasets, respectively.

\begin{table}
\centering
 \resizebox{1\columnwidth}{!}{
\begin{tabular}{lccc}
  \toprule[0.4mm]
 \rowcolor{color3}  \textbf{Methods}  & \textbf{Imbalanced} &  \multicolumn{2}{c}{\textbf{Performances}}  \\
   \rowcolor{color3}    & \textbf{Split} &  Exp. 1 (5-classes) & Exp. 2 (10-classes) \\
  \midrule
  Hierarchical-KNN \cite{ballerini2013color} & \cmark & {74.3 $\pm$ 2.5\%}  & 68.8 $\pm$ 2.0\%\\
  Hierarchical-Bayes \cite{ballerini2012non} & \cmark & {69.6 $\pm$ 0.4\%} & 63.1 $\pm$ 0.6\% \\
  Flat-KNN \cite{ballerini2013color} & \cmark & {69.8 $\pm$ 1.6\%} & 64.0 $\pm$ 1.3\%\\
  \midrule
Baseline CNN \cite{khan2017cost}  & \cmark &  75.2 $\pm$ 2.7\% & 69.5 $\pm$ 2.3\% \\
  CoSen CNN \cite{khan2017cost}  & \cmark & \sbest{80.2 $\pm$ 2.5\%} & \sbest{72.6 $\pm$ 1.6\%}  \\
 This paper & \cmark & \best{\textbf{95.7 $\pm$ 1.2}\%} & \best{\textbf{86.9 $\pm$ 0.7}\%} \\ 
  \bottomrule[0.4mm]
\end{tabular}
}\vspace{-0.5em}
\caption{Experimental results for Skin Lesion Classification on the DIL dataset.}\vspace{-0.2cm}
\label{tab:DILexp}
\end{table}

\begin{table}
\centering
 \resizebox{1\columnwidth}{!}{
\begin{tabular}{lcc}
  \toprule[0.4mm]
  \rowcolor{color3}
  \textbf{Methods}  & \textbf{Imbalanced Split} &  \textbf{Performances}  \\
  \midrule
  Deeply Supervised Nets \cite{lee2015deeply} & \xmark & 99.6\%\\
  Generalized Pooling Func. \cite{lee2016generalizing} & \xmark & \sbest{99.7\%} \\
  Maxout NIN \cite{chang2015batch} & \xmark & \best{99.8\%} \\
  \midrule
  Baseline CNN \cite{khan2017cost} & \cmark & 97.1\% \\
  CoSen CNN \cite{khan2017cost} & \cmark & \sbest{98.4\%} \\
  This paper & \cmark & \best{98.7\%} \\
  \bottomrule[0.4mm]
\end{tabular}
}\vspace{-0.5em}
\caption{Results for digit classification on MNIST.}\vspace{-0.2cm}
\label{tab:MNISTexp}
\end{table}

\begin{table}
\centering
 \resizebox{1\columnwidth}{!}{
\begin{tabular}{p{30mm} >{\columncolor{color4}}p{2.5mm} p{2mm} p{2mm} p{2mm} p{3.5mm} | p{2mm} p{2mm} p{2mm} p{2mm} p{2mm} p{5mm}}
\toprule[0.4mm]
\rowcolor{color3}
\textbf{Attributes} & \rotatebox{90}{Imbalance level} & \rotatebox{90}{Triplet-kNN \cite{schroff2015facenet}} & \rotatebox{90}{Panda \cite{zhang2014panda}} & \rotatebox{90}{ANet \cite{liu2015faceattributes}} & \rotatebox{90}{DeepID2 \cite{sun2014deep}} & \rotatebox{90}{Over-sampling \cite{drummondc4}} & \rotatebox{90}{Under-sampling \cite{drummondc4}} & \rotatebox{90}{Threshold Adj. \cite{chen2006decision}}  & \rotatebox{90}{Cost-sensitive \cite{he2009learning}} & \rotatebox{90}{LMLE \cite{huang2016learning}} & \rotatebox{90}{\textbf{This paper}}  \\
\midrule
Attractive & 1  & 83 & 85 & \sbest{87} & 78 & 77 & 78 & 69 & 78 & \best{88} & 77 \\
Mouth Open & 2 & 92 & \sbest{93} & \best{96} & 89 & 89 & 87 & 89 & 89 & \best{96} & \sbest{93} \\
Smiling & 2 & 92 & \sbest{98} & 97 & 89 & 90 & 90 & 88 & 90  & \best{99} & 90 \\
Wear Lipstick & 3 & 91 & \sbest{97} & 95 & 92 & 92 & 91 & 89 & 91 & \best{99} & 94 \\
High Cheekbones & 5 & 86 & 89 & \sbest{89} & 84 & 84 & 80 & 83 & 85 & \best{92} & 83 \\
Male & 8 & 91 & \best{99} & \best{99} & 94 & 95 & 90 & 95 & 93 & \best{99} & \sbest{98} \\
Heavy Makeup & 11 & 88 & 95 & \sbest{96} & 88 & 87 & 89 & 89 & 89 & \best{98} & 91 \\
Wavy Hair & 18 & 77 & 78 & 81 & 73 & 70 & 70 & 77 & 75 & \sbest{83} & \best{84} \\
Oval Face & 22 & 61 & 66 & 67 & 63 & 63 & 58 & \best{72} & 64 & \sbest{68} & 67 \\
Pointy Nose & 22 & 61 & 67 & \sbest{69} & 66 & 67 & 63 & \best{72} & 65 & \best{72} & \best{72} \\
Arched Eyebrows & 23 & 73 & 77 & 76 & 77 & \sbest{79} & 70 & 76 & 78 & \sbest{79} & \best{83} \\
Black Hair & 26 & 82 & 84 & \sbest{90} & 83 & 84 & 80 & 86 & 85 & \best{92} & \sbest{90} \\
Big Lips & 26 & 55 & 56 & 57 & 62 & 61 & 61 & \sbest{66} & 61 & 60 & \best{69} \\
Big Nose & 27 & 68 & 72 & 78 & 73 & 73 & 76 & 76 & 74 & \best{80} & \sbest{79} \\
Young & 28 & 75 & 78 & \sbest{84} & 76 & 75 & 80  & 24 & 75 & \best{87} & 72 \\
Straight Hair & 29 & 63 & 66 & 69 & 65 & 66 & 61 & \sbest{73} & 67 & \sbest{73} & \best{81} \\
Brown Hair & 30 & 76 & 85 & 83 & 79 & 82 & 76 & 81 & 84 & \best{87} & \sbest{86} \\
Bags Under Eyes & 30 & 63 & 67 & 70 & 74 & 73 & 71 & \sbest{76} & 74 & 73 & \best{83} \\
Wear Earrings & 31 & 69 & 77 & \sbest{83} & 75 & 76 & 70 & 76 & 76 & \sbest{83} & \best{89} \\
No Beard & 33 & 82 & 87 & \sbest{93} & 88 & 88 & 88 & 15 & 88 & \best{96} & 85 \\
Bangs & 35 & 81 & 92 & 90 & 91 & 90 & 88 & 93 & 90  & \best{98} & \sbest{95} \\
Blond Hair & 35 & 81 & 91 & 90 & 90 & 90 & 85 & 92 & 89 & \best{99} & \sbest{95} \\
Bushy Eyebrows & 36 & 68 & 74 & 82  & 78 & 80 & 75 & \sbest{84} & 79 & 82 & \best{87} \\
Wear Necklace & 38 & 50 & 51 & 59 & 70 & \sbest{71} & 66 & 62 & \sbest{71} & 59 & \best{76} \\
Narrow Eyes & 38 & 47 & 51 & 57 & 64 & 65 & 61 & \sbest{71} & 65 & 59 & \best{75} \\
5 o'clock Shadow & 39 & 66 & 76 & 81 & \sbest{85} & \sbest{85} & 82 & 82 & 84 & 82 & \best{91} \\
Receding Hairline & 42 & 60 & 67 & 70 & 81 & 82 & 79 & \sbest{83} & 81 & 76 & \best{86} \\ 
Wear Necktie & 43  & 73 & 85 & 79 & 83 & 79 & 80 & 76 & 82 & \sbest{90} & \best{92} \\
Eyeglasses & 44 & 82 & 88 & 95 & 92 & 91 & 85 & 95 & 91 & \sbest{98} & \best{99} \\
Rosy Cheeks & 44 & 64 & 68 & 76 & 86 & \sbest{90} & 82 & 82 & \best{92} & 78 & 89 \\
Goatee & 44 & 73 & 84 & 86 & 90 & 89 & 85 & 89 & 86 & \sbest{95} & \best{96} \\
Chubby & 44 & 64 & 65 & 70 & 81 & \sbest{83} & 78 & 81 & 82 & 79 & \best{87}  \\
Sideburns & 44 & 71 & 81 & 79 & 89 & \sbest{90} & 80 & 89 & \sbest{90} & 88 & \best{95} \\
Blurry & 45 & 43 & 50 & 56 & 74  & 76 & 68 & \sbest{78} & 76 & 59 & \best{87} \\
Wear Hat & 45 & 84 & 90 & 90 & 90 & 89 & 90 & 95 & 90 & \best{99} & \sbest{97} \\
Double Chin & 45 & 60 & 64 & 68 & 83 & \sbest{84} & 80 & 83 & \sbest{84} & 74 & \best{85} \\
Pale Skin & 46 & 63 & 69 & 77 & 81 & 82 & 78 & \sbest{85} & 80 & 80 & \best{89} \\
Gray Hair & 46 & 72 & 79 & 85 & 90 & 90 & 88 & \sbest{91} & 90 & \sbest{91} & \best{94} \\
Mustache & 46 & 57 & 63 & 61 & 88 & \sbest{90} & 60 & 86 & 88 & 73 & \best{91} \\
Bald  & 48 & 75 & 74 & 73 & \sbest{93} & 92 & 79 & \sbest{93} & \sbest{93} & 90 & \best{95} \\
\midrule
Overall & -  & 72 & 77 & 80 & 81 & 82 & 78 & 79 & 82 & \sbest{84} & \best{87} \\
\bottomrule[0.4mm]
\end{tabular}}\vspace{-0.5em}
\caption{Multi-label attribute prediction results on CelebA dataset. The compared methods are divided into two categories (a) \emph{left:} methods without class imbalance learning and (b) \emph{right:} methods that focus on imbalance learning.}\vspace{-0.5em}
\label{tab:celebA}
\end{table}

\begin{SCtable*}[][!htb]
\centering
 \resizebox{.77\textwidth}{!}{
\begin{tabular}{lcccccc}
\toprule[0.4mm]
\rowcolor{color3}
{\textbf{Methods}} & \multicolumn{6}{c}{\textbf{Performances}}      \\ \cline{2-7}
\rowcolor{color3}           &  Accuracy  & Precision & Recall & F1 & G-Mean & IBA \\ \toprule[0.4mm]

Cost-sen SVM \cite{cao2013optimized}           &   34.2   &  60.9$\pm$33.6 & 34.3$\pm$32.4 & 30.5$\pm$23.0 & 69.7$\pm$29.4 & 55.5$\pm$30.6   \\
NearMiss \cite{mani2003knn} & 63.5 & 66.1$\pm$14.9 & 63.5$\pm$19.7 & 62.8$\pm$14.2 & 79.2$\pm$9.5 & 61.8$\pm$14.8 \\
SMOTE NN \cite{chawla2002smote}                  & 76.3  & \sbest{80.5$\pm$12.6} & 76.3$\pm$ 22.1 & 75.0$\pm$12.3 & \sbest{88.3$\pm$6.4} & \sbest{77.1$\pm$11.9} \\
ADASYN NN \cite{he2008adasyn}                &  {76.4} & 80.4$\pm$12.2 & {76.4$\pm$21.1} & 75.2$\pm$10.6 & 88.2$\pm$6.0 & 76.9$\pm$11.4 \\
Under-samp. Clustering \cite{yen2009cluster} & 75.7 & 78.3$\pm$12.2 & 75.7$\pm$10.3 & \sbest{75.9$\pm$6.9} & 87.0$\pm$6.8 & 74.8$\pm$12.2 \\
Neighborhood Cleaning \cite{laurikkala2001improving} & \sbest{76.7}	& 79.2$\pm$10.4 &	\sbest{76.7$\pm$18.6} & 75.9$\pm$8.6 & 85.7$\pm$10.2	& 73.2$\pm$18.1 \\
Instance Hardness \cite{smith2014instance} & 67.8 & 76.3$\pm$19.8 &	67.8$\pm$18.7 & 	67.6$\pm$10.8	& 79.9$\pm$11.1	& 63.4$\pm$17.4 \\
This Paper                   & \best{80.6}    & \best{80.8$\pm$6.1}	& \best{80.6$\pm$9.3} &	\best{80.4$\pm$6.3}	& \best{88.7$\pm$5.2}  &	\best{77.6$\pm$9.6}	 \\
\bottomrule[0.4mm]
\end{tabular}}\hspace{-0.8em}
\caption{Performance on imbalanced CIFAR-10. Standard deviation on class-specific performance is reported for each metric. Our method performs better compared to others on a diverse set of evaluation metrics.\vspace{-4em}}\vspace{-1em}
\label{res: cifar10}
\end{SCtable*}

\noindent \textbf{Skin Lesion Detection:} The results for skin lesion detection are listed in Table~\ref{tab:DILexp}. We perform a 3-fold cross validation to report the results. Compared to a recent cost-sensitive CNN approach \cite{khan2017cost}, we obtain an impressive absolute performance gain of $15.5\%$ and $14.3\%$ on Experiments 1 (5 classes) and 2 (10 classes), respectively.

\noindent \textbf{Digit Classification:} For hand-written digit classification, we report our results on an imbalanced split of MNIST dataset in Table~\ref{tab:MNISTexp}. For the sake of comparison, we also report some representative methods on the original balanced split of MNIST. However, the two set of techniques are not directly comparable since the bottom group uses $\sim 45\%$ less training data. Our technique outperforms other imbalance learning approaches.

\begin{figure}
\centering
\includegraphics[clip,trim=0cm 0cm 8.5cm 0cm,width=\columnwidth]{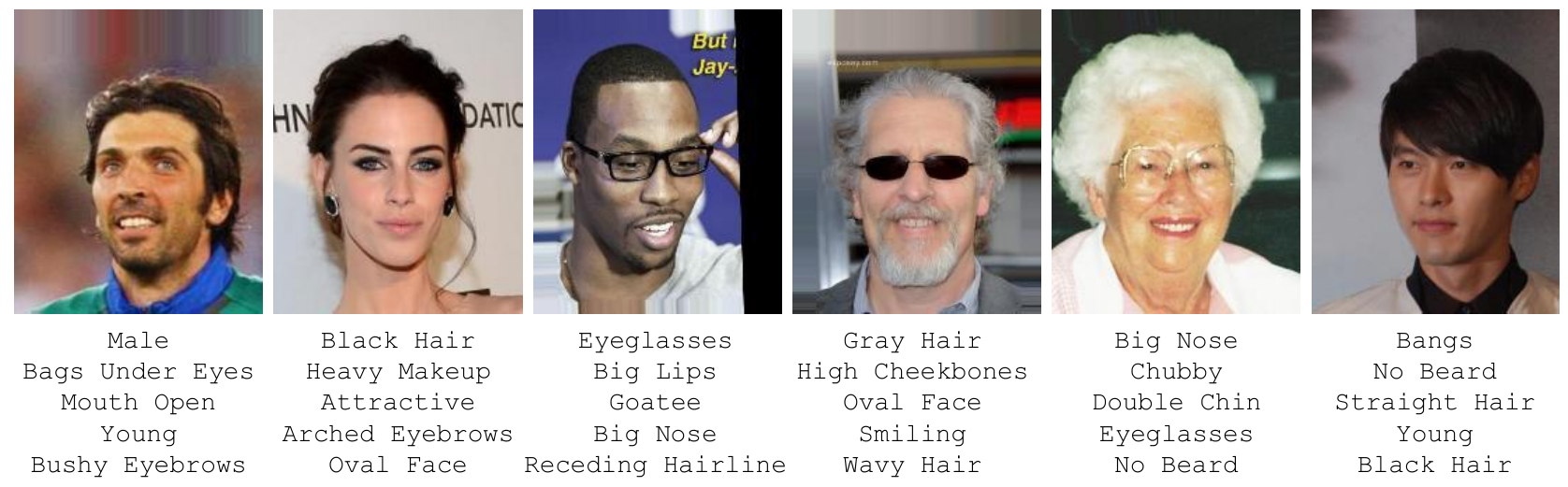}
\vspace{-1.6em}
\caption{Sample attribute predictions on the CelebA.}
\label{fig:attribute}
\end{figure}

\noindent \textbf{Attribute Prediction:}  We report the multi-label attribute prediction results on CelebA dataset in Table~\ref{tab:celebA}. This dataset is particularly challenging as it exhibits a high imbalance with majority-to-minority ratio up to $1{:}43$. We compare with nine recent state-of-the-art methods.  These methods include techniques that specifically focus on class imbalance learning (Table~\ref{tab:celebA}, right block). Our approach performs significantly better compared to both normal and class imbalance learning methods. Specifically, we achieve top-most accuracy in 23/40 attributes and second-best accuracy in other 8/40 classes. In particular, since our method focuses on assigning a larger classification region to under-represented classes, we achieve more pronounced boost for the case of highly imbalanced classes. For example, out of the top $50\%$ most imbalanced classes (with fewer samples), we achieve best performance in $16/20$ ($80\%$) cases. Qualitative examples for attribute prediction are shown in Fig.~\ref{fig:attribute}.

\noindent \textbf{Object Classification:} We compare against a number of popular imbalanced learning approaches including both data-level (e.g., SMOTE NN \cite{chawla2002smote}, ADASYS NN \cite{he2008adasyn}) and algorithm-level imbalance removal techniques (e.g., cost-sensitive SVM \cite{cao2013optimized}, NearMiss \cite{mani2003knn}), on imbalanced CIFAR-10 dataset (by retaining only $10\%$ of the samples for $50\%$ classes). Table~\ref{res: cifar10} summarizes our results in terms of a diverse range of metrics (e.g., accuracy, F1, G-mean, IBA, precision, recall). These metrics provide a comprehensive view on the performances of ours and other imbalanced techniques and are more suitable for imbalanced learning scenarios. Our results show that the proposed uncertainty based technique consistently performs better than the traditional imbalance removal methods. Particularly, for the case of evaluation metrics that give equal importance to rare classes (e.g., recall and F1 measure), our approach achieves significant performance boost of $3.9$ and $4.5$, respectively. Furthermore, the deviation of individual class scores is much lower compared to other techniques.

\noindent \textbf{Ablation Study:}
We experiment with different variants of our approach in Table~\ref{table:ablation}. Specifically, we report the performance of a simple baseline model with softmax loss and compare it with the uncertainty-based margin (UMM) enforcement and sample-level uncertainty modeling (SUM). We note a progression in performance from UMM to UMM+SUM. This is because the SUM penalizes hard examples which further helps in improving generalizability. We also study the effect of changing the dropout rate on our uncertainty-based approach. Increasing the dropout rate beyond 0.5 does not help, while a smaller rate of 0.2 or 0.3 results in a performance drop. 

\begin{SCtable}
\centering
\captionsetup{justification=raggedright,singlelinecheck=false}
\resizebox{.75\columnwidth}{!}{
\begin{tabular}{l c  c c }
\toprule[0.4mm]
\rowcolor{color3}
\textbf{Method} ($\downarrow$) &   \multicolumn{3}{c}{\textbf{Accuracy}}  \\
\midrule
CNN + Softmax loss & \multicolumn{3}{c}{97.2} \\
\midrule
\texttt{Dropout Rate} ($\rightarrow$) & \texttt{0.3} & \texttt{0.5} & \texttt{0.7} \\
\midrule
CNN + UMM loss & 98.0 & 98.3 & 98.2 \\
CNN + UMM + SUM (Ours) & 98.3 & 98.7 & 98.7  \\
\bottomrule[0.4mm]
\end{tabular}
}\hspace{-0.7em}
\caption{Ablation study on imbalanced MNIST dataset. }
\label{table:ablation}
\end{SCtable}

\noindent \textbf{Other margin enforcing losses with uncertainty:} We also experiment with recent variants of softmax loss that explicitly enforce margin constraints during classification along with uncertainty estimates (Table \ref{table:softmax variants}). These methods include Additive Margin Softmax \cite{wang2018additive}, Angular (Arc) Margin Softmax \cite{deng2018arcface}, Large Margin Softmax \cite{liu2016large} and SphereFace \cite{liu2017sphereface}. As these loss functions have generally been proposed for face verification, we already compare with them in Table~\ref{tab:lfw_ytf}. However, here our main goal is to analyze if the uncertainty estimates help in learning better boundaries for the rare classes. To this end, we use exactly the same features for all techniques and use uncertainty estimates in place of their original parameter $m$ settings. Since our application is different from face verification, we note a relatively lower performance from SphereFace and a higher performance from Large Margin Softmax. Overall, the uncertainty scores help in achieving discriminativeness between difficult classes and gives better performance in all cases.

\begin{table}
\centering
\resizebox{1\columnwidth}{!}{
\begin{tabular}{l c c c c}
\toprule[0.4mm]
\rowcolor{color3} \textbf{Loss Type} & \multicolumn{2}{c}{\textbf{Settings}} & \textbf{Original} & \textbf{with Uncer.} \\
\cline{2-3}
\rowcolor{color3} & $n$ & $m$ & & \\
\midrule[0.4mm]
Softmax (Baseline) & - & - & 76.8 & - \\
Additive Margin \cite{wang2018additive} & 30 & 0.35 & 78.3 & 78.9 \\
Arc Margin \cite{deng2018arcface} & 30 & 0.4 & 78.1 & 78.4 \\
Large Margin \cite{liu2016large} & - & 4 & 79.4 & 79.9 \\
Sphere Product \cite{liu2017sphereface}  & - & 4 & 76.2 & 77.5 \\
\bottomrule[0.4mm]
\end{tabular}
}\vspace{-0.5em}
\caption{The behavior of other recent margin-based loss formulations with uncertainty. The accuracy values are reported for imbalanced CIFAR-10 dataset. \emph{`n'}, \emph{`m'} stands for feature norm and margin, respectively.}
\label{table:softmax variants}
\end{table}

\section{Conclusion}
We present a new approach to address class imbalance problem, underpinned by the Bayesian uncertainty estimates. We demonstrate that the classifier confidence levels are directly associated with: (a) the difficulty-level of individual samples and (b) the scarcity of the training data for under-represented classes. Our proposed approach utilizes uncertainty to enforce larger classification regions for rare classes and challenging training samples.  This results in better generalization  of learned classifier to new samples for less frequent classes. We achieve significant performance gains on several datasets for face verification, attribute prediction, object/digit classification and skin lesion detection. 

{\small
\bibliographystyle{ieee}
\bibliography{DB}
}

\end{document}